# Deep Learning Advances in Vision-Based Traffic Accident Anticipation: A Comprehensive Review of Methods, Datasets, and Future Directions

Ruonan Lin,[1,^] Tao Tang, [4,^] Yongtai Liu,[3,^] Wenye Zhou,[1] Xin Yang,[1] Hao Zheng,[1] Jianpu Lin[1,2,*] and Yi Zhang,[1,*]
[1]School of Advanced Manufacturing, Fuzhou University, Quanzhou 362251, China.
[2]National Local United Engineering Lab of Flat Panel Display Technology, Fuzhou University, Fuzhou 350116, China.
[3]College of computer software, Hanyang University, Seoul 04763, Republic of Korea.
[4]College of Computing, City University of Hong Kong, Hong Kong 999077, China.
[^]These authors contributed equally
*Correspondence: jianpu.lin@fzu.edu.cn (J.P. Lin)

## SUMMARY

Traffic accident prediction and detection are critical for enhancing road safety, and vision-based traffic accident anticipation (Vision-TAA) has emerged as a promising approach in the era of deep learning. This paper reviews 147 recent studies, focusing on the application of supervised, unsupervised, and hybrid deep learning models for accident prediction, alongside the use of real-world and synthetic datasets. Current methodologies are categorized into four key approaches: image and video feature-based prediction, spatio-temporal feature-based prediction, scene understanding, and multimodal data fusion. While these methods demonstrate significant potential, challenges such as data scarcity, limited generalization to complex scenarios, and real-time performance constraints remain prevalent. This review highlights opportunities for future research, including the integration of multimodal data fusion, self-supervised learning, and Transformer-based architectures to enhance prediction accuracy and scalability. By synthesizing existing advancements and identifying critical gaps, this paper provides a foundational reference for developing robust and adaptive Vision-TAA systems, contributing to road safety and traffic management.

## INTRODUCTION

The introduction does not include a heading and should expand on the background of the topic, typically including in-text citations1.Traffic accidents continue to pose a significant global issue, leading to substantial personal and economic losses each year. The United Nations has set a goal to halve the number of road traffic deaths by 2030 and has designated 2021-2030 as the Decade of Action for Road Safety1. According to the latest report from the World Health Organization, approximately 1.19 million people lose their lives in road traffic accidents globally each year2. Consequently, developing effective traffic accident detection and prediction technologies remains critically important.

Traditional methods for traffic accident detection and prediction have primarily relied on machine learning techniques such as decision trees and support vector machines3,4. While these approaches perform well on small-scale datasets, they frequently fail to capture complex patterns and dynamic changes in large-scale, high-dimensional traffic data, thereby restricting their predictive capabilities5. However, the emergence of deep learning presents novel opportunities for automation and enhancing predictive accuracy6.

Fang et al.7pioneered the concept of vision-based traffic accident anticipation (Vision-TAA), providing a comprehensive review of primary research approaches, challenges, and benchmarks, thus establishing the groundwork for the first Vision-TAA survey. Iqbal H. Sarker8offered an extensive overview of deep learning techniques, while Noushin Behboudi et al.9undertook a comprehensive review of the most recent advances in machine learning and deep learning for traffic accident analysis and prediction over the past five years, focusing on risk, frequency, severity, and duration prediction.

Compared to traditional machine and rule-based approaches, deep learning excels in feature extraction, representation, and processing of vast and complex datasets. Its end-to-end learning capabilities, adaptability, and generalization have rendered it a focal point of research in recent years10,11,12,13. Furthermore, with the increasing availability of video data from surveillance cameras and dashcams, the potential for developing and optimizing Vision-TAA systems has expanded significantly.

In this study, we reviewed 147 papers published from previous years through 2024, providing a comprehensive overview of the state of development and research advancements in deep learning for traffic accident prediction. This review covers the various deep learning methods applied in this field and offers an in-depth analysis and synthesis of these approaches.

## RESULTS

In the field of traffic accident prediction, the selection of datasets is critical for evaluating and optimizing algorithms. To provide a comprehensive review of Vision-TAA research, we will begin by detailing some of the most commonly used datasets.

### Real-World Traffic Scene Datasets

These datasets are derived from real traffic scenarios, including dashcam and surveillance camera footage, and are primarily used for traffic accident detection, risk assessment, and driving behavior analysis.

- KITTI[14]
  This dataset encompasses a variety of driving environments, including urban, rural, and highway settings.
- CCD[15]
  The CCD dataset provides annotations on environmental attributes and accident causes, and all recorded accidents involve vehicle collisions.
- CADP[16]
  The CADP dataset consists of 230 videos, each containing at least one accident captured by a fixed surveillance camera, along with 1,416 traffic accident clips. Among these, 205 clips include complete spatiotemporal annotations.
- DAD[17]
  This dataset contains accident videos from 678 online users across six cities. The accidents are manually annotated with the time, location, and moving objects. The video types are distributed as follows: 42.6% involve motorcycle-car collisions, 19.7% involve car-to-car collisions, 15.6% involve motorcycle-motorcycle collisions, and the remaining 20% involve other types of incidents.
- Near Accident Database (NIDB)[18]
  The NIDB contains over 62,000 video clips and 1.3 million frames, most of which depict various event scenarios. The video content is categorized into four major object classes: cyclists, pedestrians, vehicles, and background (negative) examples.
- Pedestrian Near-Miss (PNM)[19]

  This dataset captures tense moments before and after accidents, with annotations assessing the associated risks. Pedestrian risk levels are categorized as high, low, or no risk.
- A3D[20]
  Similar to the DAD dataset, A3D focuses on accident initiation, pinpointing the accident's start at the 80th frame in each sequence. This dataset includes 1,500 video clips from East Asian regions.

### Multi-Task Risk Prediction and Driving Behavior Analysis Datasets

These datasets focus on multi-task prediction, driving risk assessment, and accident prediction, widely applied in driving behavior analysis.

- DRAMA[21]
  The DRAMA dataset includes 17,785 driving interaction scenarios conducted in Tokyo. It addresses object-level challenges and key entities, aiming to describe visual scenarios with free-form language.
- SUTD-TrafficQA[22]
  This dataset focuses on traffic event reasoning by analyzing 10,080 real-world scene videos and 62,535 question-answer pairs. It presents six challenging tasks designed to evaluate system understanding and reasoning about complex traffic events.
- DADA-2000[23]
  The DADA-2000 dataset is a new benchmark containing 2,000 video sequences, totaling 658,476 frames, and includes annotations for driver attention and accident intervals.

### Simulated and Synthetic Datasets

These datasets are derived from simulated or synthetic environments and are commonly used to test and train accident prediction models in controlled settings, where replicating real accident scenarios is unsafe.

- GTACrash[24]
  The GTACrash dataset, extracted from the Grand Theft Auto V video game, includes 3,661 non-accident scenes and 7,720 accident scenes.
- DoTA[25]
  This dataset comprises 4,677 videos from two YouTube channels, annotated with detailed temporal, spatial, and categorical information. It classifies accidents into nine types based on the movement of vehicles, pedestrians, and other obstacles, with vehicles representing 73.2% of the dataset.

### Deep Learning Models

Deep learning models are generally categorized into supervised and unsupervised learning. In this domain, hybrid models are frequently employed to address the complexities of traffic scenarios. This section introduces several common models used in Vision-TAA.

#### *Supervised Deep Networks*

- **Multilayer Perceptron (MLP):** The Multilayer Perceptron (MLP)[26] is a fundamental supervised learning algorithm commonly applied in traffic accident prediction for feature learning[27] , object localization[21], and data fusion[27]. It falls under the category of feedforward artificial neural networks (ANN) and serves as the foundation for building deep learning models.

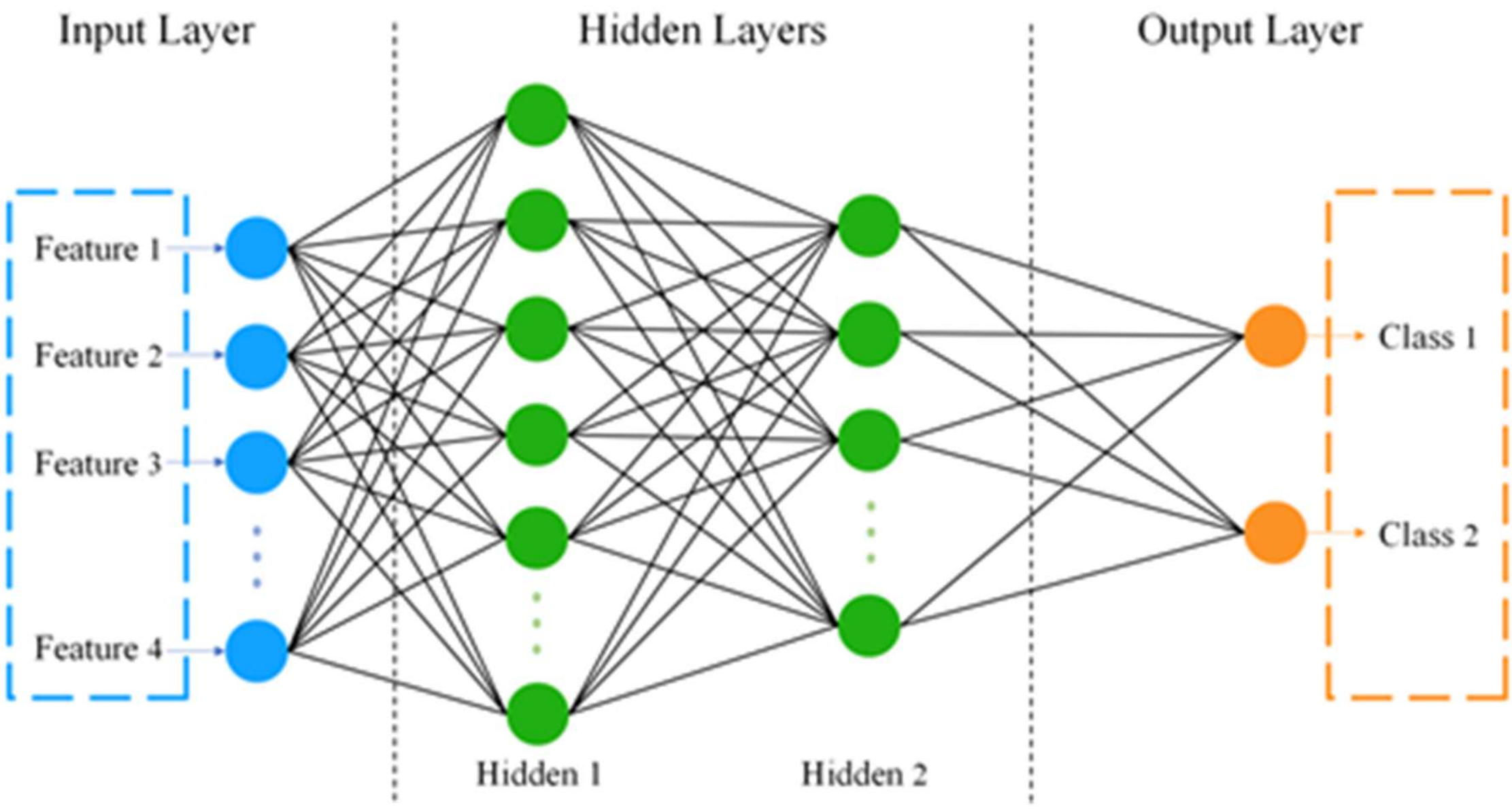

Figure. 1. Multilayer Perceptron

- **Convolutional Neural Networks (CNN)**: Convolutional Neural Networks (CNN)[28] effectively capture local features and patterns from images through convolution operations, preserving spatial hierarchies from low-level features to high-level semantic features. In Vision-TAA systems, CNNs are used to extract spatial features from video frames and analyze sequential frames to detect accident indicators. For instance, 3D-CNN captures spatiotemporal information from video streams to predict potential traffic accidents29,30.

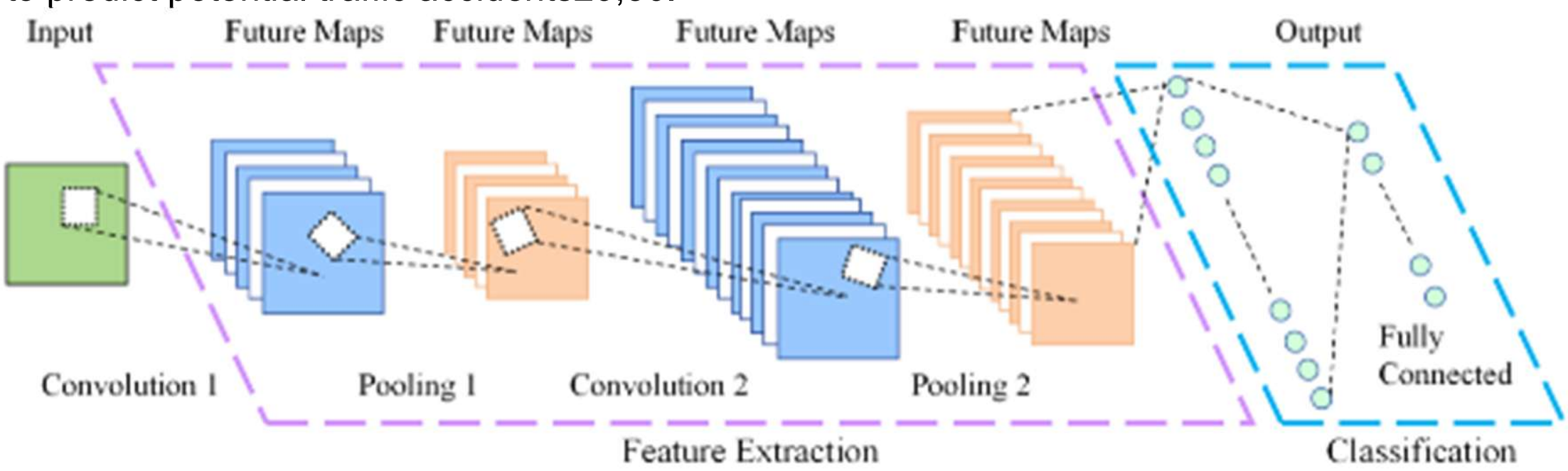

Figure. 2. Convolutional Neural Networks

- **Recurrent Neural Networks (RNN) and Variants**: Unlike CNNs, Recurrent Neural Networks (RNN) [31] are well-suited for handling sequential data by utilizing feedback connections to retain historical information. However, standard RNNs suffer from the vanishing gradient problem, limiting their capacity to process long sequences. Variants such as Long Short-Term Memory (LSTM) and Gated Recurrent Units (GRU) mitigate this issue through gating mechanisms that improve gradient flow and performance[32]. In Vision-TAA, RNNs and their variants are crucial for analyzing the temporal dependencies in video data[17,33,34].

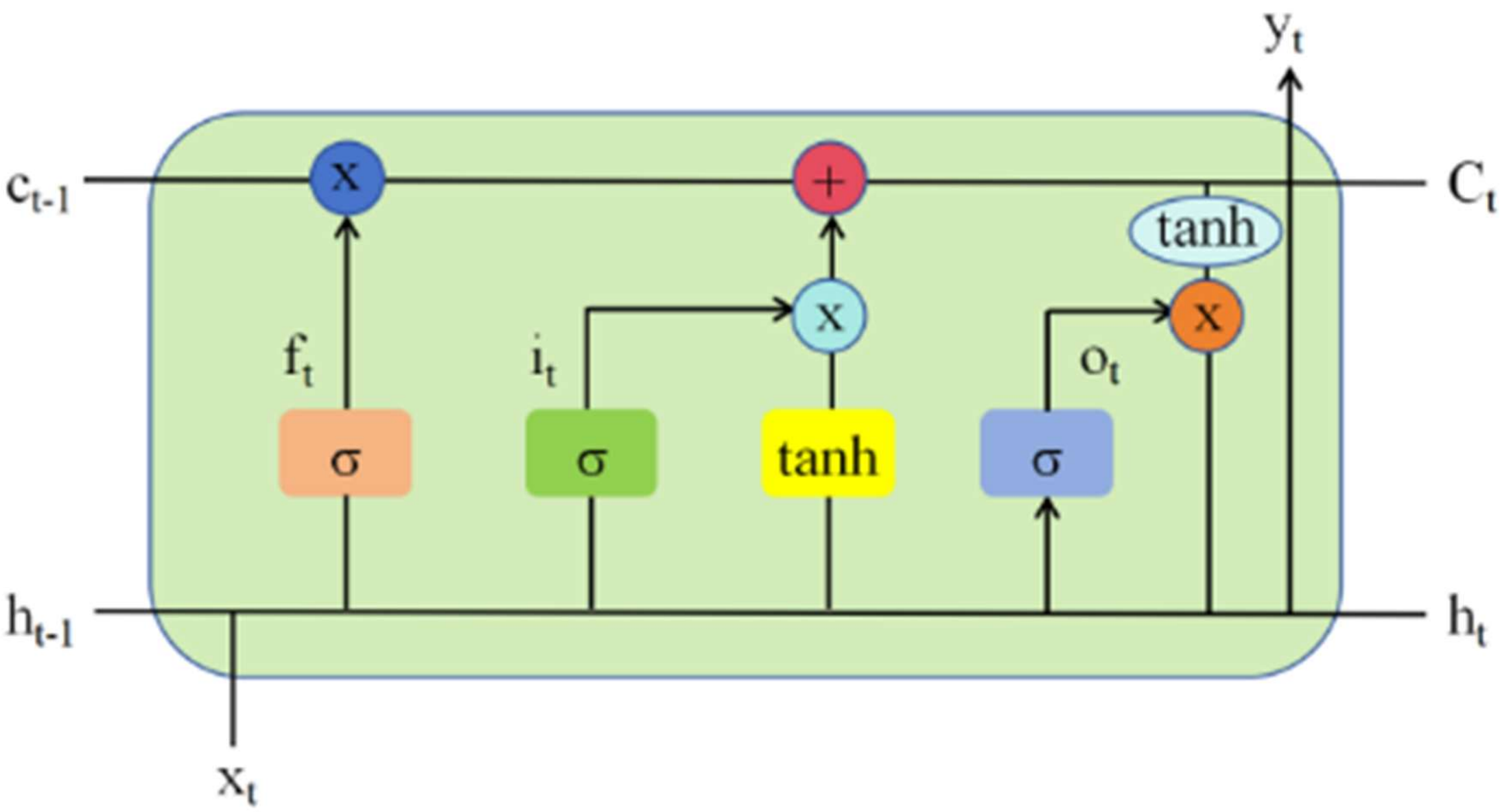

**Figure. 3. Recurrent Neural Networks and Variants**

***Unsupervised Learning Networks***

- **Generative Adversarial Networks (GAN)**: Generative Adversarial Networks (GAN)[35], introduced by Ian Goodfellow, are an unsupervised learning approach where two networks, the generator (G) and the discriminator (D), compete against each other. GANs are employed to generate synthetic traffic accident videos, enhancing the robustness and adaptability of accident prediction models[36,37,38].

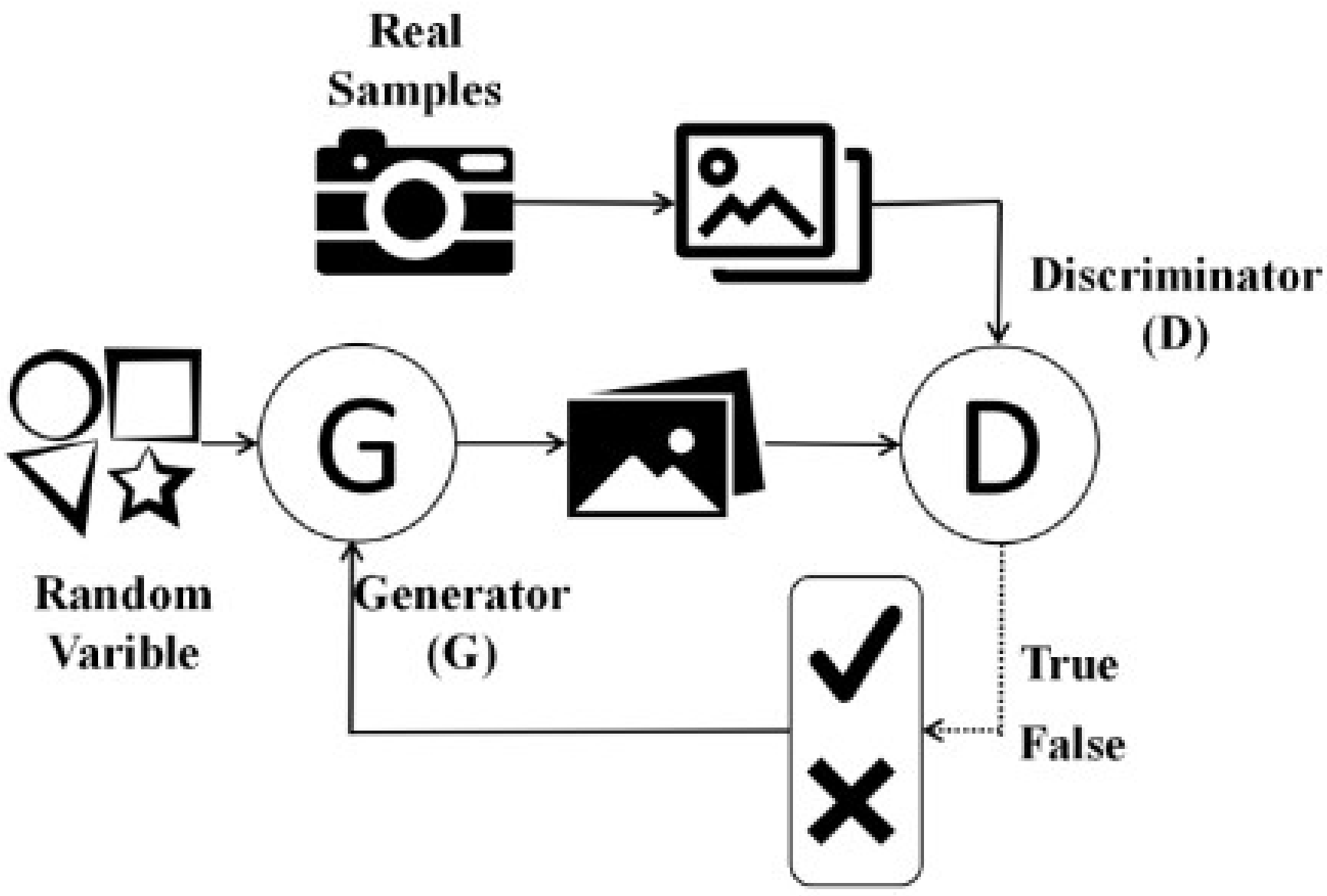

**Figure. 4. Generative Adversarial Networks**

- **Transformer**: The Transformer is a deep learning architecture based on attention mechanisms. Unlike GANs, Transformers use self-attention and multi-head attention techniques to process sequential data and capture relationships between elements, while supporting parallel computation. In Vision-TAA, Transformers, combined with CNNs, analyze video data to identify critical accident patterns, improving prediction accuracy and supporting traffic safety analysis and prevention[39,40].

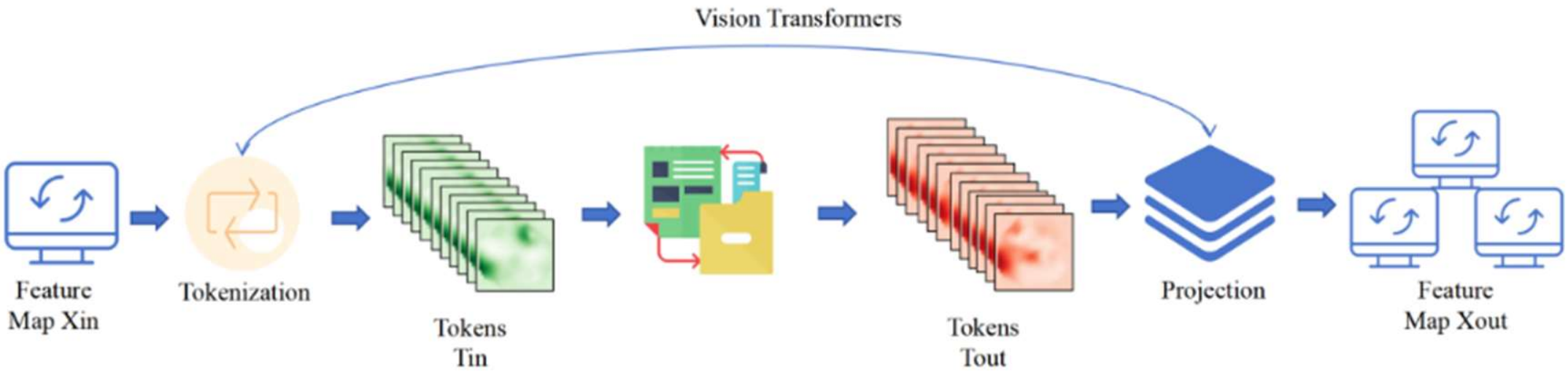

**Figure. 5. Transformer**

- **Graph Neural Networks (GNN)**: Graph Neural Networks (GNN)[41] play a critical role in handling graph-structured data, essential for tasks such as action recognition in computer vision. GNNs utilize a message-passing mechanism between nodes to capture relationships and global information flow across the graph. In Vision-TAA, GNNs simulate spatial interactions between traffic participants, such as vehicles and pedestrians[12,42].

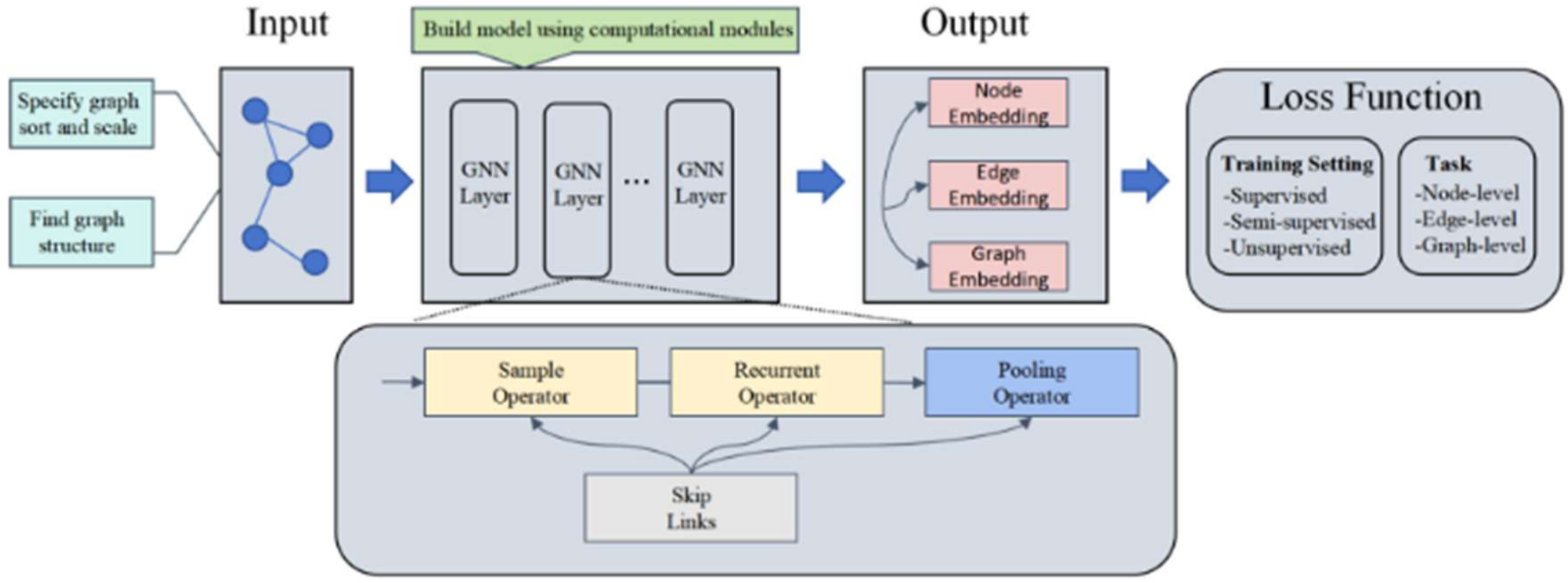

**Figure. 6. Graph Neural Networks**

***Hybrid Deep Neural Networks***

- **Single-Stage Object Detection Techniques: SSD and YOLO**: Single Shot MultiBox Detector (SSD) [43] is an object detection framework that detects multiple object candidates and their classes in a single forward pass through the network. In Vision-TAA systems, SSD is used for real-time monitoring of traffic participants such as vehicles and pedestrians, providing data for accident prediction[19]. YOL [44] divides the input image into grids, and each grid is responsible for detecting objects within its boundaries, predicting both position and class. YOLO is used for frame-level detection of vehicles and pedestrians in video streams, combined with temporal analysis to gain deeper insights and predict potential traffic accidents[45,46].

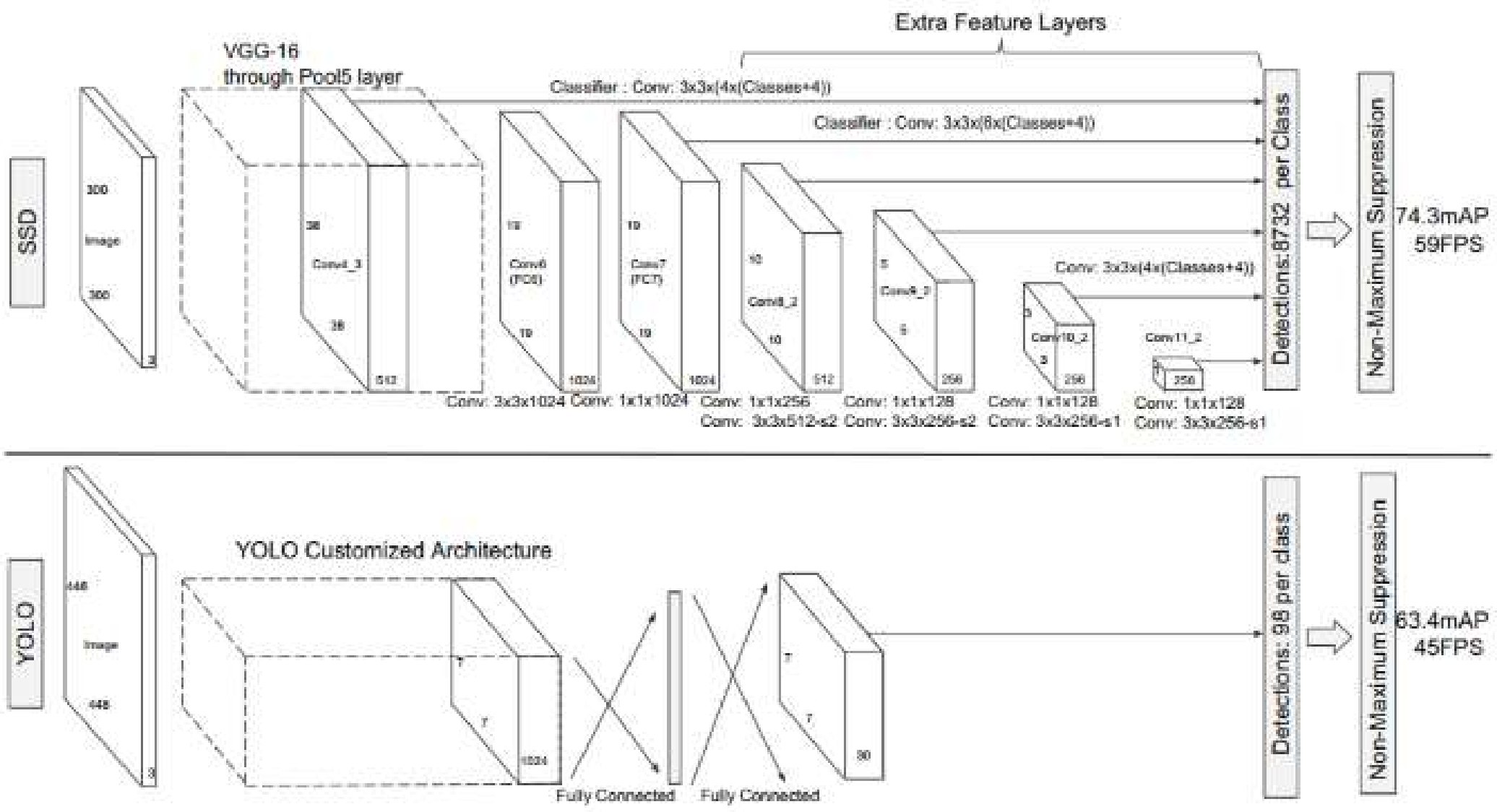

Figure. 7. SSD and YOLO

- **Region-Based Convolutional Neural Networks (R-CNN)**: Region-Based Convolutional Neural Networks (R-CNN)[47] is a two-stage deep learning object detection technique that predicts object categories and bounding boxes by first identifying image regions. Faster R-CNN allows for rapid detection of vehicle and pedestrian locations, which, when combined with temporal models, can predict accidents[13,48].

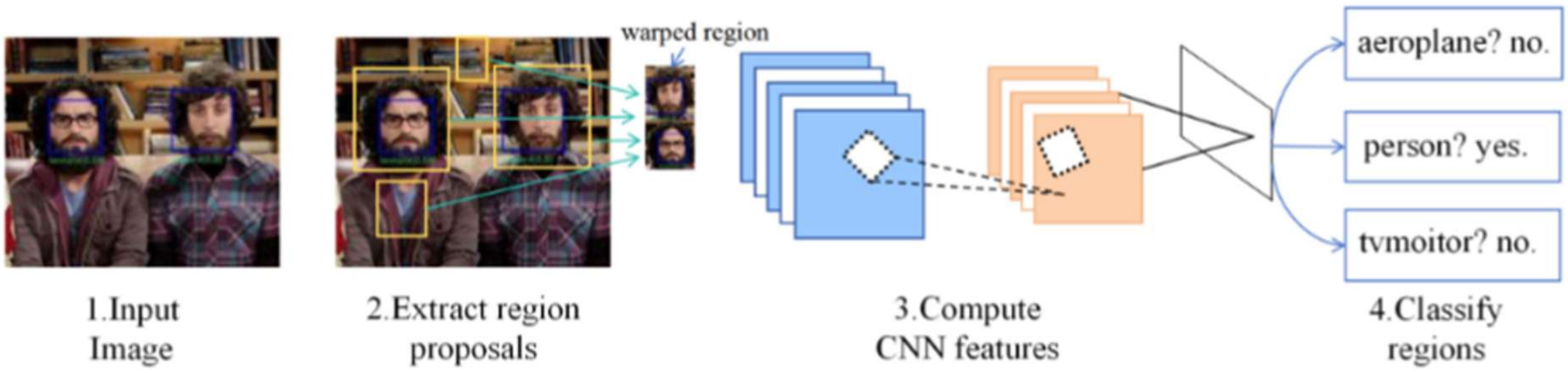

Figure. 8. Region-Based Convolutional Neural Networks

Scholars have widely applied the previously mentioned benchmarks and deep learning models in Vision-TAA. For instance, Muhammad Monjurul Karim et al.[49] utilized bounding boxes and optical flow features to identify traffic participants involved in accidents through attention modules. Wentao Bao et al.[50] enhanced the decision-making process by simulating both bottom-up and top-down attention mechanisms, which improved the visual interpretability of the model. Jianwu Fang et al.[51] used human-inspired cognitive mechanisms, combining textual descriptions with driver attention to facilitate model training. Additionally, Yu Li et al.[52] proposed a dynamic spatio-temporal attention network (DSTA) to predict the probability of future accidents.

Based on the literature from the past five years, we have identified trends and classified the applications of deep learning in Vision-TAA into four categories: image and video feature-based prediction, spatiotemporal feature-based prediction, scene understanding-based prediction, and multimodal data fusion-based prediction. Each category represents distinct research perspectives and methodologies, complementing one another to collectively advance the Vision-TAA field.

## Image and Video Feature-Based Prediction

Image and video feature-based prediction methods primarily include convolutional neural networks (CNN), Single Shot Multibox Detector (SSD), and YOLO, among other image processing techniques. By extracting spatial features from images and video frames, these methods can detect and classify objects and their attributes (e.g., vehicles, pedestrians, road signs).

For instance, Fu-Hsiang Chan et al.[33] proposed a Dynamic Spatial Attention (DSA) RNN to predict accidents in dashcam videos through object detection and RNN modeling. Shaoqing Ren et al.[53] introduced Faster R-CNN, which utilizes a region proposal network (RPN) to improve the efficiency of object detection and achieve state-of-the-art accuracy on multiple datasets. Qian Liu (2024) et al.[54] innovated Faster R-CNN by developing the Mask R-CNN model, which automatically identifies road surface voids and defects in ground-penetrating radar (GPR) images. Similarly, Marwa Chacha Andrea et al.[55] used the YOLOv8 model trained on a large set of images extracted from YouTube, achieving an average accuracy of 0.954 in accident detection. Basheer Ahmed et al.[45] proposed a method that combines the YOLOv5 object detector with the DeepSORT tracker for vehicle detection and tracking, utilizing parallel computing techniques to reduce system complexity and inference time.

While image and video feature-based prediction methods leverage deep learning's powerful visual feature extraction capabilities to analyze raw images and videos, they often rely on short-time fragments of data and struggle with the continuity of complex traffic scenarios. Additionally, these methods tend to require significant computational resources, leading to suboptimal real-time performance, particularly in handling complex backgrounds and occlusions. In certain scenarios, researchers prefer spatio-temporal feature-based prediction methods to address these limitations.

### Spatio-Temporal Feature-Based Prediction

Spatio-temporal feature-based methods capture the ped a novel model called GSNet, which learns spatiotemporal correlations in both geographical and semdynamic changes of traffic participants across time and space, making them suitable for predicting the timing and location of traffic accidents. These methods emphasize the continuity and dependency of time-series data. Models like Long Short-Term Memory (LSTM) and Recurrent Neural Networks (RNN) excel at processing temporal data.

Spatio-temporal features not only include spatial information such as the position and speed of traffic participants but also encompass dynamic behavioral patterns over time. For example, Wentao Bao et al. [20] proposed a model that combines graph convolutional networks (GCN) and recurrent networks to learn spatio-temporal relationships for predicting future accident probabilities. Tianhang Wang et al.[55] introduced a graph-based spatio-temporal continuity framework (GSC), which maintains the spatio-temporal continuity of agents through graph convolutional networks and defines a novel adjacency matrix to enhance graph learning. Zachary C et al.[53] conducted a comprehensive review and analysis of RNN research over the past 30 years, highlighting the frequent integration of RNNs with other methods in traffic accident prediction research. Yoshiaki Takimoto[56] employed convolutional recurrent neural networks (CRNN) to capture high-dimensional time-series information, achieving higher accuracy in accident prediction compared to baseline models. Wei Liu and Tao Zhang[57] proposed a new traffic accident prediction framework called THAT-Net, which integrates spatial and temporal streams to capture complementary spatiotemporal information and designed a dual-layer hidden state aggregation structure. By re-aggregating the hidden state weights of gated recurrent units (GRU) at the frame and segment levels, their method achieved the highest accident prediction accuracy, with lead times ranging from 0.48 to 2.8 seconds. Beibei Wang et al.[58] develoantic dimensions for traffic accident risk prediction. Muhammad Monjurul Karim et al.[49] proposed a dynamic spatio-temporal attention network (DSTA) that anticipates accidents from dashcam videos by combining a dynamic time attention (DTA) module and a dynamic spatial attention (DSA) module, jointly trained with a GRU to predict the probability of future accidents. Their model outperformed state-of-the-art methods on two benchmark datasets.

Unlike image and video feature-based prediction methods, spatio-temporal feature-based methods introduce a temporal dimension to capture the dynamic characteristics of traffic scenarios over time. While these methods effectively address the continuity limitations of image-based models, they may struggle with long time-series data and require large amounts of labeled temporal data, which increases the complexity of data acquisition and processing. Additionally, these methods are sensitive to noise and anomalies, which can interfere with predictions. To better address the challenges of complex traffic scenarios, researchers have further developed scene understanding-based prediction methods.

### Scene Understanding-Based Prediction

This category emphasizes a comprehensive understanding of traffic scenes and multi-target analysis by integrating various environmental information and background knowledge. It is particularly applicable to complex traffic environments and the interaction analysis of multiple traffic participants. Early research in scene understanding-based prediction mainly focused on the surveillance domain, using direct physical distance calculations for accident prediction[49,59,60,61,62]. In recent years, the focus has shifted toward deep learning, with Huang et al. [63] proposing a gated graph convolutional multi-task model that combines urban road images and traffic accident data to achieve higher prediction accuracy. Jewel Rana Palit et al.[6] proposed a hybrid model—a convolutional gated recurrent neural network (CNN + GRU). Zhou et al.[64] proposed the RiskOracle framework, which uses a multi-task differential time-varying graph network and hierarchical region selection to significantly improve prediction performance. Muhammad Haris et al.[65] proposed a computer vision-based vehicle collision prediction process that tracks vehicles under CCTV views and uses Gaussian distribution approximations to predict vehicle trajectories by leveraging the short-term stationarity of time processes.

Scene understanding-based prediction methods focus on the deep understanding of objects and their relationships within traffic scenes. This approach goes beyond merely recognizing explicit visual features and involves the extraction of semantic information. These methods excel at capturing organized information in complex traffic scenes, but they face the challenge of requiring large amounts of labeled data and sophisticated model architectures to accurately capture and interpret semantic information. Moreover, the generalization ability of these models to different environments is limited. To overcome the limitations of single-mode data, researchers have advanced multimodal data fusion-based prediction methods.

### Multimodal Data Fusion-Based Prediction

Multimodal data fusion-based prediction methods combine information from different data sources, such as video, sensor data, and weather data, to conduct comprehensive analysis and predictions. This approach compensates for the shortcomings of relying on a single data source, improving both the accuracy and robustness of predictions.

For example, Ji-Won Baek and Kyungyong Chung[11] proposed a novel data collection and analysis framework that consistently gathers diverse information through modality fusion. In a study by Camilo Gutierrez-Osorio et al.[44], they demonstrated that combining two or more machine learning techniques yields superior results, emphasizing the importance of incorporating heterogeneous data sources into models and predictions. Wentao Bao et al. proposed the DRIVE model, which simulates human visual attention mechanisms to improve accident prediction performance. Bao et al.[50] proposed a deep reinforcement accident prediction model that incorporates driver attention within the accident window. Hilmil Pradana[66] presented a method that combines visual attention behaviors and object tracking to learn and predict the current state of the driver on the road while analyzing other moving objects. Jianwu Fang et al.[23] designed a semantic context-guided attention fusion network (SCAFNet), which integrates semantic context features of images and models these features using a graph convolutional network (GCN).

By integrating data from various sensors, such as visual, radar, and LiDAR data, this method facilitates a more comprehensive understanding of traffic scenes. The advantage of this approach lies in its ability to significantly improve prediction accuracy and robustness, though it also faces challenges related to data fusion complexity and the coordination between different data modalities. Researchers need to develop novel algorithmic designs to address these issues.

Lastly, for ease of reference, Table 1 summarizes the attributes of deep learning methods applied in Vision-TAA.

**Table 1. Attributes of Deep Learning Methods Applied in Vision-TAA.**

| Author | year | model | input | cue | accuracy |
|---|---|---|---|---|---|
| Chan, F et al. [17] | 2017 | dynamic-Spatial-attention RNN | Dashcam | object interaction | 74.35% |
| Zeng, K. H et al. [67] | 2017 | RNN | human accident video<br>Dashcam | track interaction | 75.1%<br>51.4% |
| Suzuki, T et al. [18] | 2018 | Quasi-RNN<br>AdaLEA | Dashcam | Frame Risk | 99.1%<br>62.1% |
| Bao, W et al. [15] | 2020 | GCN, CNN<br>Bayesian neural network | Dashcam | object feature relationship | 53.70% |
| Formosa, N et al. [13] | 2020 | Regional CNN, DNN | Front Camera | object interaction | 67.50% |
| Kataoka, H et al. [19] | 2020 | SSD- mr model | Dashcam | object risk | 72.12% |
| Karim, M. M et al. [39] | 2021 | Gated circulation unit network | Dashcam | object feature relationship | 94.02% |
| Karim, M. M et al. [49] | 2022 | Dynamic spatiotemporal attention network<br>Gated circulation unit network | Dashcam | Frame level hidden description | 72.60% |
| Wang, T et al. [20] | 2023 | GCN | Dashcam | object interaction | 60.4%<br>94.9% |
| Liu, W et al. [57] | 2023 | Hyber model,GRU | Dashcam | Frame processing | 77.8%<br>99.5% |
| Basheer Ahmed, M et al. [45] | 2023 | DeepSORT tracker's YOLOv5 target detector | Live CCTV traffic monitoring video stream | object tracking | 99.20% |
| Lee, H et al. [68] | 2023 | Lightweight object detection model | Intersection camera | object risk | 20%<br>59% |
| LL Li et al. [69] | 2024 | Transformation fusion<br>BERT model<br>Self-attention model<br>Gated circulation unit network | Dashcam<br>Driver attention chart<br>textual description | Frame level hidden description | 74.4% |
| Thakur, N et al. [70] | 2024 | GCN | Dashcam | Frame processing | 63.6%<br>99.9% |
| Liao, H et al. [27] | 2024 | Attention mechanism<br>Large language model<br>Multimodal large-scale model | Dashcam | object feature | 69.2%<br>99.7%<br>96.4% |
| Song, W et al. 42 | 2024 | GCN<br>Attention mechanism | Dashcam | Frame processing | 75.2%<br>79.1% |
| Dao, M et al. 46 | 2024 | Deephoff YOLOv8<br>Lstm | Dashcam<br>Weather sensor | Object distance | 96.01% |

## METHODS

The methods section must provide sufficient information for the reader to be able to reproduce the study. Contents must adhere to all applicable journal policies. When further details are placed in the supplemental information, the following text should be stated in the methods section: "Further details regarding the methods can be found in the supplemental information."

### *Data Collection and Preprocessing*

This study encompasses 147 recent papers spanning from 2018 to 2024. The datasets used are detailed below:

- Real-world datasets like KITTI and CADP offer dashcam and surveillance footage for accident detection.
- Simulated datasets such as GTAC rash provide synthetic data for safe model training.

Data preprocessing involves frame extraction, annotation organization, data augmentation via GANs, and dataset division.

### *Model Training and Evaluation*

Model training utilizes frameworks like TensorFlow and PyTorch with GPU acceleration. Hyperparameters are optimized via grid search and cross-validation. The evaluation metrics include accuracy, precision, recall, and F1 score.

Further details regarding the methods can be found in the supplemental information.

## DISCUSSION

Our comprehensive survey highlights the essential role of vision-based traffic accident detection and anticipation (Vision-TAA) in enhancing road safety. However, Vision-TAA continues to encounter various challenges, including data scarcity, imbalance, and the complexity and heterogeneity of traffic environments, all of which significantly constrain model performance. To effectively address these challenges, novel approaches are imperative—particularly multimodal data fusion techniques, which can improve the robustness and interpretability of prediction models. Such advancements are crucial not only for optimizing model performance but also for fostering public trust in autonomous systems.

In recent years, deep learning technologies have made remarkable progress, with Generative Adversarial Networks (GANs) showing great promise in addressing data scarcity. Deep Convolutional GANs (DCGANs) have been highly effective in generating realistic synthetic data, balancing datasets, and improving detection accuracy. Furthermore, attention mechanisms such as YOLO and Faster R-CNN have enhanced object detection and tracking in complex traffic scenarios. Self-supervised learning is also emerging as a powerful tool for traffic accident prediction, leveraging vast amounts of unlabeled data and significantly reducing the reliance on costly annotations.

Simultaneously, the development of Vision Transformers has opened up novel possibilities for enhancing traffic accident prediction. These technological advancements, combined with multimodal data fusion, will provide a solid foundation for Vision-TAA, advancing the field toward more efficient and safer road traffic management. Future research should continue to explore these key areas to overcome existing technical barriers and promote breakthrough developments in the practical application of Vision-TAA.

## RESOURCE AVAILABILITY

### *Lead contact*

For further information, contact Jianpu Lin (jianpu.lin@fzu.edu.cn).